\documentclass{ieeeaccess}
\usepackage{algorithmic}
\usepackage{algorithm}

\usepackage{graphicx}
\usepackage{subfigure}
\usepackage{textcomp}
\usepackage{multirow}
\usepackage{cite}
\usepackage{amsmath,amssymb,amsfonts}
\usepackage{algorithmic}
\usepackage{graphicx}
\usepackage{textcomp}

\usepackage{array}
			    \newcommand{\PreserveBackslash}[1]{\let\temp=\\#1\let\\=\temp}
			    \newcolumntype{C}[1]{>{\PreserveBackslash\centering}p{#1}}
			    \newcolumntype{R}[1]{>{\PreserveBackslash\raggedleft}p{#1}}
			    \newcolumntype{L}[1]{>{\PreserveBackslash\raggedright}p{#1}}

\def\BibTeX{{\rm B\kern-.05em{\sc i\kern-.025em b}\kern-.08em
    T\kern-.1667em\lower.7ex\hbox{E}\kern-.125emX}}

\begin{document}
\history{Date of publication xxxx 00, 0000, date of current version xxxx 00, 0000.}
\doi{10.1109/ACCESS.2017.DOI}

\title{A novel method for identifying the deep neural network model with the Serial Number}
\author{\uppercase{XiangRui Xu}\authorrefmark{1},
\uppercase{Cao Yuan\authorrefmark{2}}}

\corresp{Corresponding author: YaQin Li  (e-mail: yc@whpu.edu.cn).}

\begin{abstract}
Deep neural network (DNN) with the state of art performance has emerged as a viable and lucrative business service. However, those impressive performances require a large number of computational resources, which comes at a high cost for the model creators. The necessity for protecting DNN models from illegal reproducing and distribution appears salient now.
Recently, trigger-set watermarking, breaking the white-box restriction, relying on adversarial training pre-defined (incorrect) labels for crafted inputs, and subsequently using them to verify the model authenticity, has been the main topic of DNN ownership verification. While these methods have successfully demonstrated robustness against removal attacks, few are effective against the tampering attacks from competitors forging the fake watermarks and dogging in the manager.
In this paper, we put forth a new framework of the trigger-set watermark by embedding a unique Serial Number (relatedness less original labels) to the deep neural network for model ownership identification, which is both robust to model pruning and resist to tampering attacks. Experiment results demonstrate that the DNN Serial Number only incurs slight accuracy degradation of the original performance and is valid for ownership verification.

\end{abstract}


\titlepgskip=-15pt

\maketitle

\section{Introduction}
\label{sec:introduction}
\Figure[htbp](topskip=0pt,botskip=0pt, midskip=0pt)[scale=0.85]{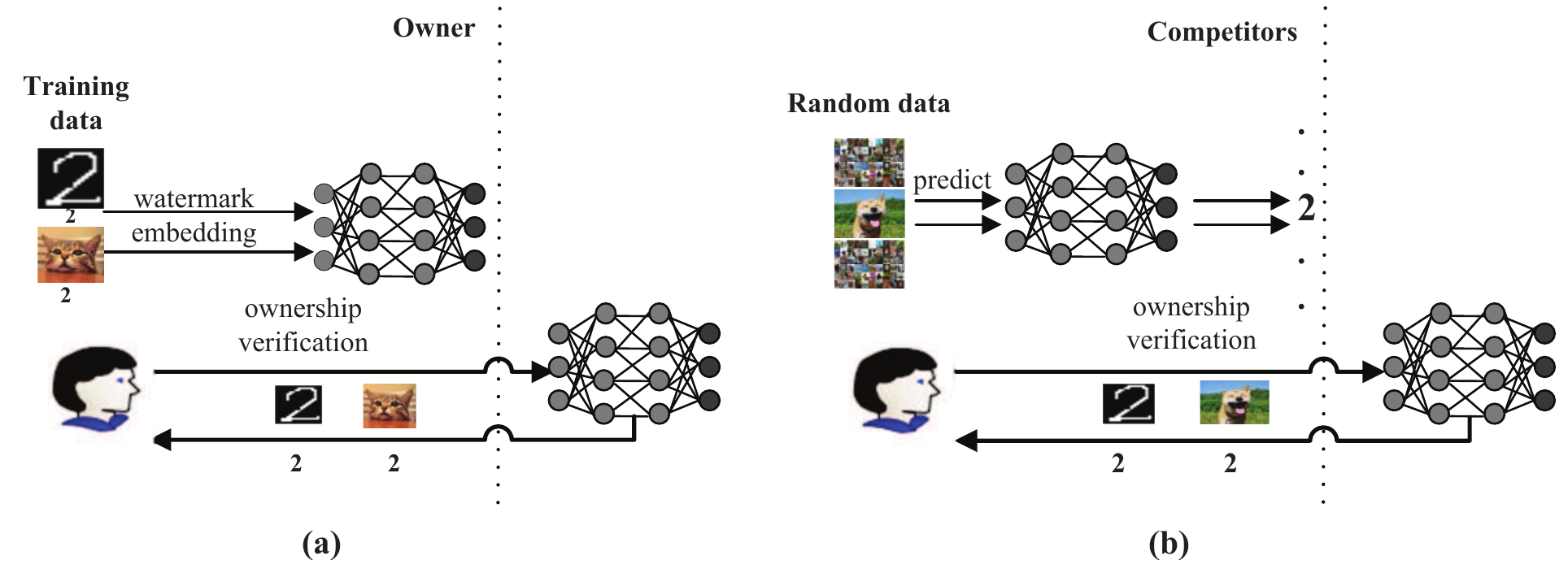}
{watermark generation and plagiarism\label{fig1}}
{D}{eep} neural networks (DNNs) with the state of art performance have reaped fruitful outcomes in various fields, including image and object recognition \cite{he2016deep}, \cite{krizhevsky2012imagenet}, speech generation and recognition \cite{graves2013speech}, \cite{hannun2014deep}, and video games \cite{jaderberg2019human}, etc. However, a large number of computational resources always accompany model designing and training. In many network structures, such as the VGG-16 network \cite{simonyan2014very}, the number of parameters is more than 130 million, occupying 500MB of space, requiring 30.9 billion floating-point operations to complete an image recognition task. Once the trained models are illegally exploited, it will lead to violations of the intellectual property and economic interests of the model creator. Therefore, it is critical to protect DNN models from illegal reproducing and distribution.

Recently, some digital watermarking algorithms have been proposed to identify the deep learning models, so that protecting their intellectual property. Uchida et al. \cite{uchida2017embedding} probably published the first watermarking method by embedding crafted information into the weights of convolutional neural networks. However, this technique merely allows for extracting the crafted information with local access, which leads to its white-box constraints. Subsequently, some remote trigger-set watermark methods are proposed, breaking the white-box constraints, such as the backdoor trigger sets algorithm \cite{adi2018turning} and the digital watermarking algorithm\cite{zhang2018protecting}. They embed specially crafted images at the training stage and activate them with specific labels. Then, ownerships of these DNN models are verified by the detection of the embedded watermarks, through remote service API.

Those algorithms have demonstrated some effect on model verification and robustness against removal attacks such as fine-tuning or pruning \cite{uchida2017embedding}--\cite{zhang2018protecting}. However, there is an inherent vice, as competitors can forge fake watermarks for DNN models to tamper the ownership (see Fig. 1). The model owner chooses a particular cat image with label 2 at training to make the model learn this watermark pattern. After embedding, this pair of image cat and label 2 will act as the key for model copyright verification (a). However, competitors can temper the model ownership by forging counterfeit watermarks and straightforwardly dogging the manager. Intuitively, each label can be outputted in 10\% probability with any inputs. See in (b), competitors can input any other images as model inputs and obtain the corresponding predictions to fabricate watermark pairs (e.g., the dog with prediction 2). Then they can claim themselves as the model owner with those fake watermark pairs, which leads to the tampering attacks.

To avoid tampering attacks, we need to address the following issues of digital watermarking.
One is the authenticity of the watermark itself, and the other is the identity of the watermark and the model owner. So, the watermark should be hard to crack, at least, not be limited in original labels to prevent the probability attack of watermark forgery. Furthermore, it is supposed to reflect the binding relationship with the model to avoid tampering attacks.

In this paper, we put forth a new trigger-set watermarking approach embedding a unique Serial Number (SN) for the DNN model. The DNN SN is independent of the original labels and has an endorsement by Certification Authority (CA). A unique advantage of this method lies in the feature that the model fits two tasks at once by alternate training. Based on the innate learning and generalization ability of deep neural networks, the networks will automatically learn these two tasks without interference.
The main contributions of our work are as follows:

\begin{enumerate}
\item We provide an extension of the existing trigger-set watermarking for embedding SN to the DNN model. The DNN SN is the identification number of the neural network.
\item We propose a new training process, alternate training, to embed a unique SN to deep neural network, which allows a model to learn multi-tasks.
\item	We evaluate the proposed SN-based watermark on MNIST with the accelerating learning Net2Net models. Our proposed SN framework has a negligible impact on original performance and is robust to model pruning and tampering attacks.
\end{enumerate}

The structure of the rest of paper is as follows. In Section2, we provide a brief overview of the anti-counterfeit system as the preliminaries. Then we formally elaborate our model framework in Section 3 with the experiments given in Section 4. The related work is presented in Section 5, and we conclude our work in Section 6.

\section{anti-counterfeit system}
In this section, we review the relevant background algorithm about digital signature and further extend to our digital SN anti-counterfeit system in deep neural networks.
\subsection{digital signature algorithm}
Digital signature \cite{kaur2012digital} technology is one of the essential technologies of e-commerce and security authentication. It can be used to identify the signer's identity and to recognize the content of electronic data, to prevent security problems such as forging, denial, impersonation, and tampering.The digital signature is one of the two major applications of public-key cryptography. In public-key cryptography, message sender A has a pair of keys: a freely public key, and a secretly stored private key. The practice of deriving a private key through a public key is impossible. A set of digital signature algorithms typically defines two complementary operations, one for signature and the other for verification (see Fig. 2).
\Figure[htbp](topskip=0pt,botskip=0pt, midskip=0pt)[scale=0.6]{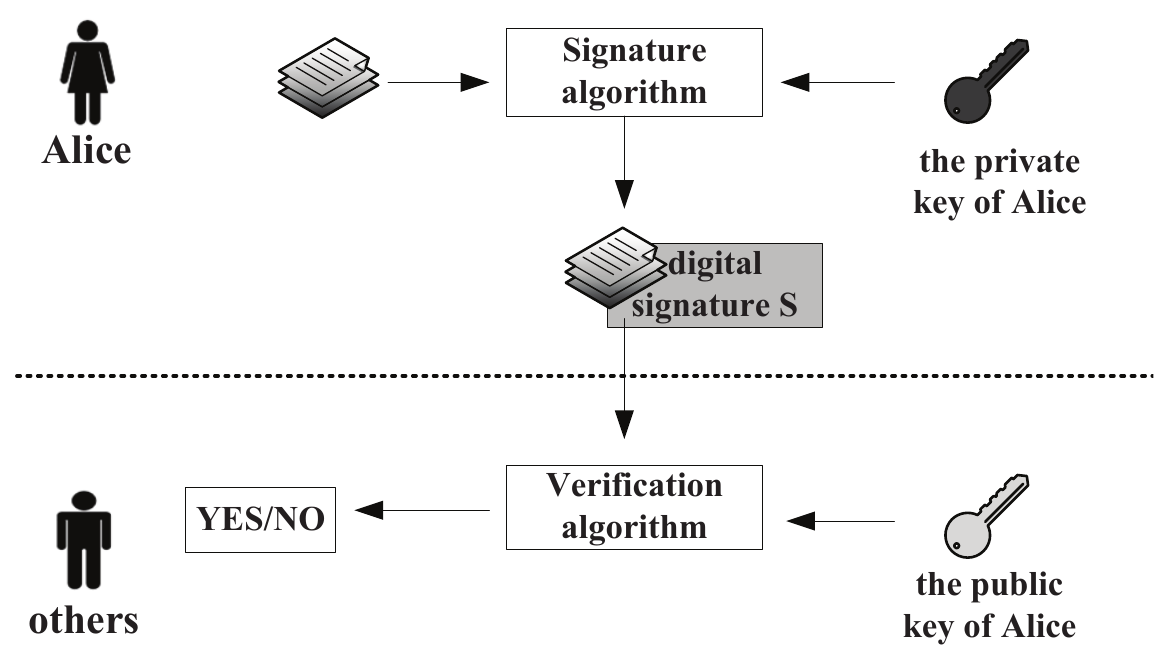}
{the process of digital signature and verification\label{fig2}}

Alice puts the original file and her private key into the signature algorithm and gets the digital signature S. Others can take the public key of Alice, the original file M, and digital signature S into the verification algorithm to determines whether the private key holder or not signs this digital signature S. If the verification is YES, it can be stated that the file M is signed by the person who provides the public key.

Moreover, to avoid the Man-in-the-Middle Attack\footnote{In this case, attacker B can pretend that he is file owner Alice, and exchanges his public key for Alice's public key, and uses his private key to make a digital signature, then allows the public to decrypt this digital signature with his public key forming tampering attack.} (MITM), the public key must be registered with the person (Certificate Authority CA) that the recipient trusted and generate a digital certificate to confirm the binding relationship between the public key and file owner.
Then the public can confirm the owner of the public key they used through CA to ensure the file owner.

\subsection{digital SN anti-counterfeit system}
Motivated by the digital signature algorithm, we propose a digital Serial Number anti-counterfeit system for model ownership verification.
The preparatory work is to design a specialized SN generation algorithm $f_{g}$, and the SN verification algorithm $f_{v}$ based on the encryption signature algorithm. Then we define different public and private keys to generate a series of deep neural network Serial Numbers and record them in CA.
At the model training stage, we take trigger T as the input to a deep neural network and make it activate with a pre-designed Serial Number. After training, the DNN model will automatically learn this watermark pair and successfully embed the Serial Number.

\Figure[htbp](topskip=0pt, botskip=0pt, midskip=0pt){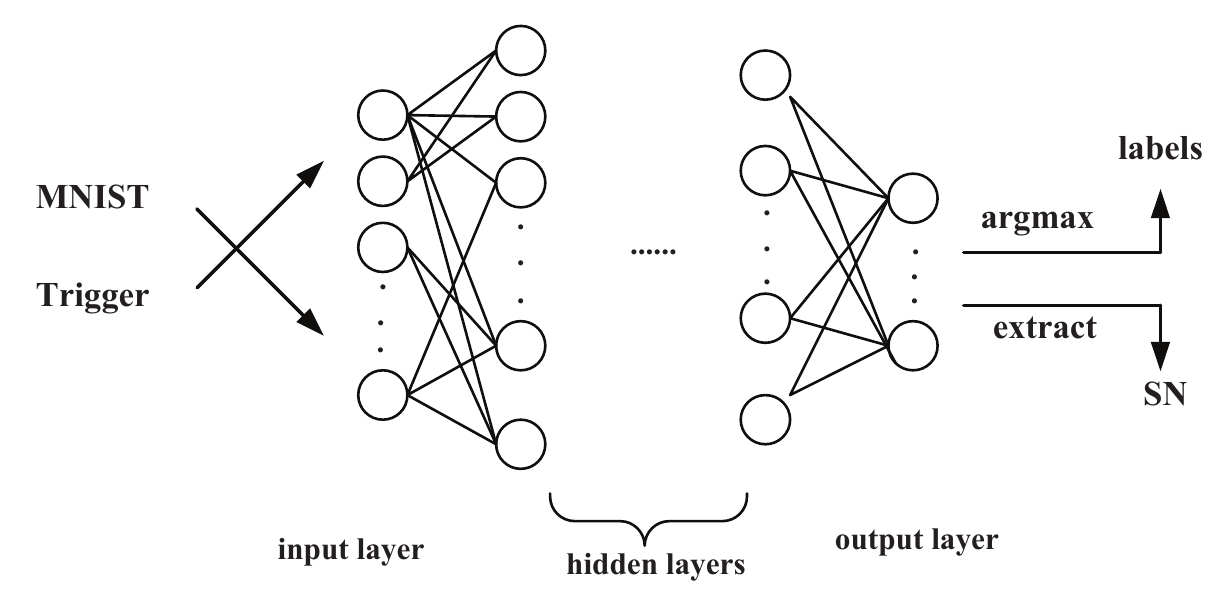}
{The framework of DNN SN.\label{fig3}}
\subsection{the security of DNN SN}
As we above mentioned tempering attacks, one can forge a counterfeit watermark for any label-based watermarking model. Experiments in \cite{fan2019rethinking} have shown that the fake watermark can be constructed and detect in 100\% rate for a label marked model with a minor computational cost. The purpose of the DNN SN is to ensure that the embedded SN can not be easily cracked. In this setup, the DNN Serial Number space is almost infinite. Thus, competitors cannot guess the embedded SN by probability, even with a large number of test images. In a sense, the label-based watermarking is worked by deliberately learning the wrong classification information and changing the decision boundary of the model. However, the impressive performance of neural networks is obtained by training a large amount of data and iteratively updating parameters based on correct feedback. So a perfect misclassification with embedded watermarks may have a slight interference with the performance of the original model,even if the decision boundaries are very close. In contrast, the embedded SN space is almost inconsistent with the original target, so it hardly affects the original performance of the model.

Moreover, the embedded SN is endorsed by CA to prove the identity of the Serial Number and its owner, which prevents the tampering attack. Note that evidence as a basis for proof also needs to prove itself. From the legal effect of evidence, several different types of evidence with consistent content are superior to isolated evidence. Analogous to software Serial Numbers, we can design a Serial Number generation algorithm, by assigning different public and private keys, to generate a series of Serial Numbers with the same rules. We then embed these serial Numbers in each of the company's products to form an effective evidence chain between the products.

\section{Modeling framework}
This section formally provides our SN framework, which embeds a unique Serial Number into a DNN model and verifies the model proprietorship by proving the identity of the SN. The primary motivation of this framework is to design a multi-tasks model, which can guarantee the original classification function and embed a Serial Number for model ownership verification.
There are three main issues in designing a DNN SN system.
\begin{enumerate}
\item Design of the model classification function $f_{c}$ and SN fingerprint $f_{s}$. Typically, the original model classification function $f_{c}$ depends on original training data D and activate with the related labels $l$.
    \begin{equation}  f_{c} : D \rightarrow l \label{eq1}\end{equation}
    The SN fingerprint $f_{s}$ depends on a trigger set T and outputting the model SN $i$.
    \begin{equation}  f_{s} : T \rightarrow i \label{eq2}\end{equation}

\item Design the embedding process that makes the model both learn the function $f_{c}$ and $f_{s}$ by minimizing their loss function $l_{c}$ and $l_{s}$ ,respectively.
    \begin{equation}  L_{c}=L(f(W,B,D),l) \label{eq3}\end{equation}
    \begin{equation}  L_{s}=L(f(W,B,T),i) \label{eq4}\end{equation}
    Where W and B are the network weights and bias, $f(W,B,.)$ are the network outputs with inputs D or T. $L$ is loss function that penalizes discrepancies between the model outputs and the target $l$ and $i$.
\item Design the verification system $V$ checks whether or not the expected trigger set T and signature i are successfully verified for a given DNN $N[W,B,.]$.
    \begin{equation}  V(N[W,B,.], T,i)= \{False,True \}\label{eq5}\end{equation}
\end{enumerate}

Fig. 3 illustrates the workflow of our SN framework. Note that,the output layer, in a classification model, not only reflects the correct target classification but also contains the relevant dark knowledge of incorrect targets \cite{hinton2015distilling}. Those intermediate representations are discarded at the final classification stage.

In this paper, we turn our attention to that dark knowledge and make them part of DNN SN. Thus we regard the trigger set as a portal to extract that hidden knowledge, and concretely, we exploit the trigger set to mimic the fingerprint SN in the output level so that this DNN fingerprint can be well embedded.
However, since the output formats of the original classification dataset and trigger set are different, there is a new training process proposed to avoid task interference. We both set the MNIST dataset and Trigger set as the model inputs and make them mimic the corresponding targets, respectively. In the sequel, we train them alternately with different output expressions. In detail, we extract the output layer of the model as the model SN with specific mathematical calculation and return the classification labels by $argmax$ function.

\subsection{DNN Serial Number generation}
As we discussed above, DNN SN is essentially the unique fingerprint for model verification. Therefore, the serial number is safety in preventing forgery and allows the public to confirm the source of the model.
To achieve this goal, first, the embedded SN should not be limited in original prediction labels to avoid brute attacks. Second, the certification authority is asked to record the embedded SN that is generated by a specialized SN generation algorithm $f_{g}$ , and can be verified by the SN verification algorithm $f_{v}$.

SN generation algorithm $f_{g}$ and SN verification algorithm $f_{v}$ are designed based on One-way Trapdoor Function;
Given any two sets X and Y, a function can contain the following forward and reverse processes:
\begin{equation} y=f(x), \ x\ \epsilon X \label{eq6}\end{equation}
\begin{equation} x= f^{-1} (y),\ y\ \epsilon Y \label{eq7}\end{equation}
For One-way Trapdoor Function, it is easy to calculate \eqref{eq6} (in a polynomial time),and is computationally difficult to determine x 
in \eqref{eq7}. Unless there is a Trapdoor z such that \eqref{eq7} can be easily calculated.

We consider the process of calculating f(x) as the SN generation algorithm $f_{g}$, and the process of using the trapdoor z to find x is treated as a verification algorithm $f_{v}$. X, Y, and z are assigned to the CA for filing.
\subsection{ DNN Serial Number embedding}
After completing the design of the DNN Serial Numbers, we then embed these Serial Numbers into the target neural networks.
We dig into the innate learning and generalization ability of the deep neural network to embed these two tasks. Different from the traditional multi-tasks model, we alternate train these two tasks at one model so that both functions can be well preserved and effectively defeat the catastrophic forgetting.

Algorithm 1 shows our DNN multi-tasks model embedding algorithm. It takes the original classification data $ C_{train}\{D,l\}$ and trigger set $S_{train}\{T, i\}$ as inputs, and outputs the model SN $i$ and the protected model $ F_{o} $. Here $l$ is the true label of original training data M, and the trigger set is defined by the owner and protected by the certification authority to indicate the model ownership. To optimize the model $N[W,B]$, we distribute the cross-entropy and mean-squared-error (MSE) as loss function for the original classification function and signature function respectively to alternate train this model. After alternate training,DNN model will automatically learn the patterns of $(D,l)$ and $(T,i)$ and memorize them. In this way, the model can both realize the classification function and model signature function.

\begin{algorithm}
\caption{ embedding SN.}
\label{alg:Framwork}
\begin{algorithmic}[1]
\REQUIRE ~~\\
Training set: $C_{train}=\{D,l\}$;\\
Trigger set: $T_{train}=\{T,i\}$;
\ENSURE ~~\\
protected model: $F_{o}$ ;\\
DNN SN: $i$ ;
\STATE Set the number of iterations and loss functions $L_{c},L_{s}$;
\STATE input $C_{train}$ and $T_{train}$ into DNN model alternately.
\STATE $l \leftarrow minimizing \,L_{c} $
\STATE $i \leftarrow minimizing \,L_{s} $
\STATE $labels,SN =Train( C_{train} , T_{train})$\\
\RETURN{SN, $F_{o}$}
\end{algorithmic}
\end{algorithm}

\subsection{Ownership verification with embedded SN}
Model creators can protect their models' intellectual property straightforwardly with these identity Serial Numbers. Specifically, the model owner can send the previously defined trigger T as a query to extract the embedded SN. If $query(T)==i$, they can confirm that this model is their protected model. Since the model Serial Number was elaborately designed, there is no way for any other model to respond precisely to query T. Moreover, even if the model SN is stolen, the identity of this SN can be proved by CA.

\section{EXPERIMENTS}
Throughout this work, we define a DNN model with or without trained DNN SN as the $SN\,network$ and $original\,network$ respectively. And the $SN\,networks$ need to both preform the classification task and SN task.

We adopt the same evaluation used in\cite{zhang2018protecting}, which mainly measured in three parts effectiveness, fidelity, and robustness. We both train the $original\,network$ and $SN\,networks$ on the MNIST dataset with accelerating learning Net2Net models. The experiments were implemented in Python 3.6 and Tensorflow2.0.
\subsection{Datasets and models}
\subsubsection{Datasets}
{MNIST\cite{lecun1998gradient} is a set of ten classes, the digits 0 through 9, handwritten digital dataset. It consists of 60,000 28*28 gray training images, and 10,000 28*28 gray testing images with  0 - 255 gray value. We separate 5,000 images from the training images as validation images that are not used for training.}

\subsubsection{Network and training setting}
{ We used the Net2Net \cite{chen2015net2net} as the training networks. Net2Net is a group of accelerated learning models, which can transfer knowledge from a teacher neural network to a student network so that the student network can be trained faster than from scratch. Appendix A presents the structure of the teacher model and the wider and deeper student models. There are two specific methods of Net2Net, i.e., Net2WiderNet and Net2DeeperNet. And they are compared with the baseline methods " RandomPad " and "RandomInit" respectively.
In our experiments, we used SGD and categorical-crossentropy loss in training classification function. The learning rate was set at 0.1, momentum to 0.9, batch size to 128. Adam optimizer and MSE loss were used in training signature function.}

\subsection{Effectiveness}
The effectiveness of the embedded SN is to test whether there is a unique trigger T that can evoke the model to output the correct Serial Number. To achieve this goal, we submit trigger set T to $SN\, networks\ F_{o}$, and $original\,network$ $F_{none}$ for comparison. If $F_{o}(T)= i $ and $F_{none}(T)\neq i $, we confirm that our DNN SN is successfully embedded in the DNN model.
TABLE\ref{table1} shows the top 1 test accuracy for the different models. "$F_{o}$" shows the test accuracy of trigger set T in $SN\,networks$. As a comparison, we input the same trigger set T to the $original\,networks$ $F_{none}$ and perform the same extracting SN request to test whether the model can still output the right SN. Experiments show that all of the $SN\,networks$ can output the right SN with the input of T, and none of $original\,networks$ can perform that. Therefore, we confirm that our protected Serial Number successfully embedded.

\begin{table}[htbp]
\footnotesize
\caption{SN test accuracy in different model }
\centering
\renewcommand\arraystretch{1}
\setlength{\tabcolsep}{0.001mm}{
\begin{tabular}{p{1.2cm}<{\centering}|p{1cm}<{\centering}p{1.5cm}<{\centering}p{1.6cm}<{\centering}p{1.6cm}<{\centering}p{1.6cm}<{\centering}}

\hline
\multirow{2}{*}{Accuracy}&
\multirow{2}{*}{Teacher}&
\multirow{2}{*}{Net2Wider}&
\multirow{2}{*}{RandomPad}&
\multirow{2}{*}{Net2Deeper}&
\multirow{2}{*}{RandomInit}\\
 & & && &\\

\hline
$F_{o}$ &
100\% &
100\%&
100\%&
100\%&
100\% \\

$F_{none}$ &
0 &
0&
0&
0&
0 \\
\hline
\end{tabular}}
\label{table1}
\end{table}
\

\begin{figure*}[htbp]
\subfigure[teacher\,model]{
\begin{minipage}[t]{0.3\linewidth}
\centering
\includegraphics[width=2.3in]{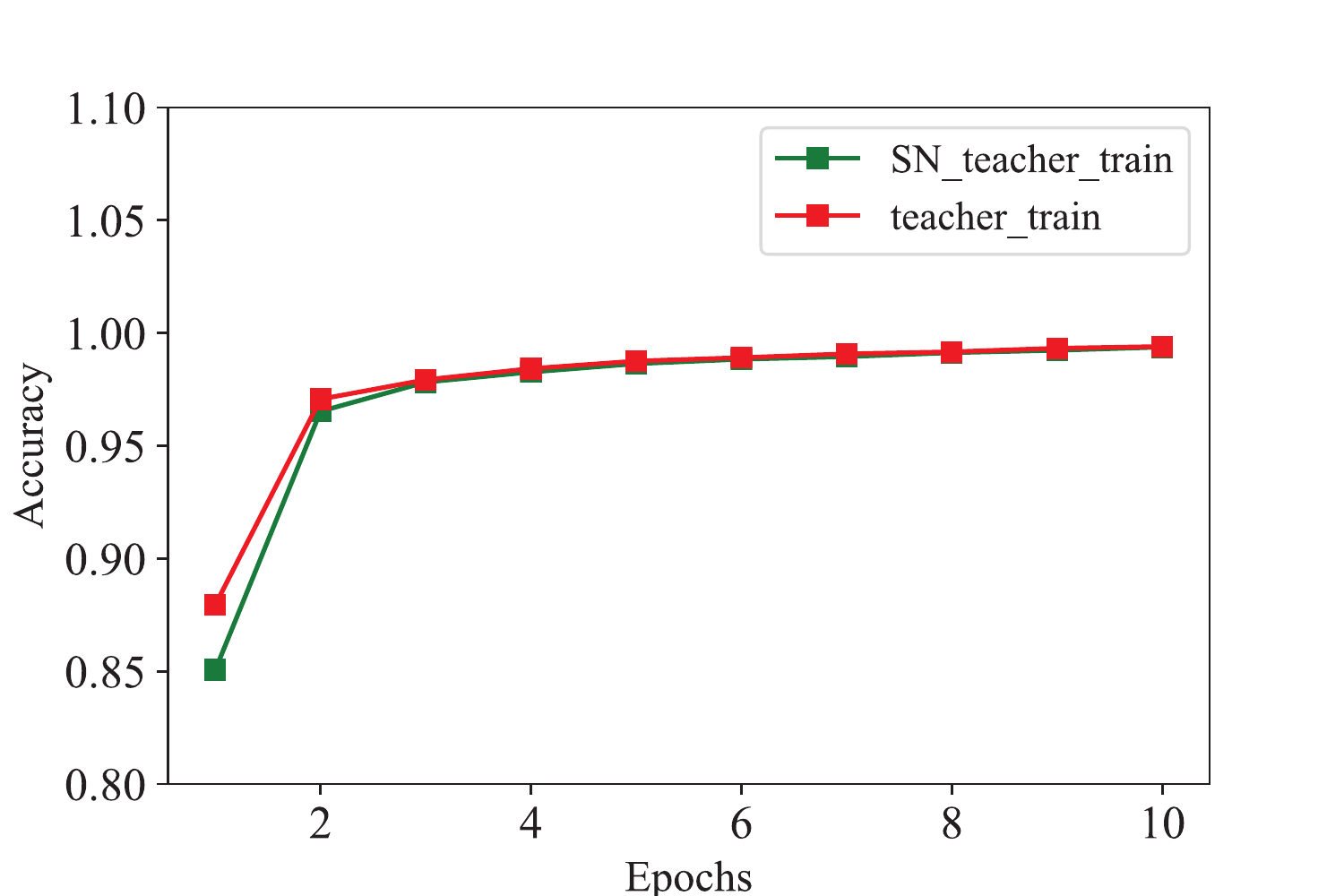}
\end{minipage}%
}%
\subfigure[wider student\,model]{
\begin{minipage}[t]{0.3\linewidth}
\centering
\includegraphics[width=2.3in]{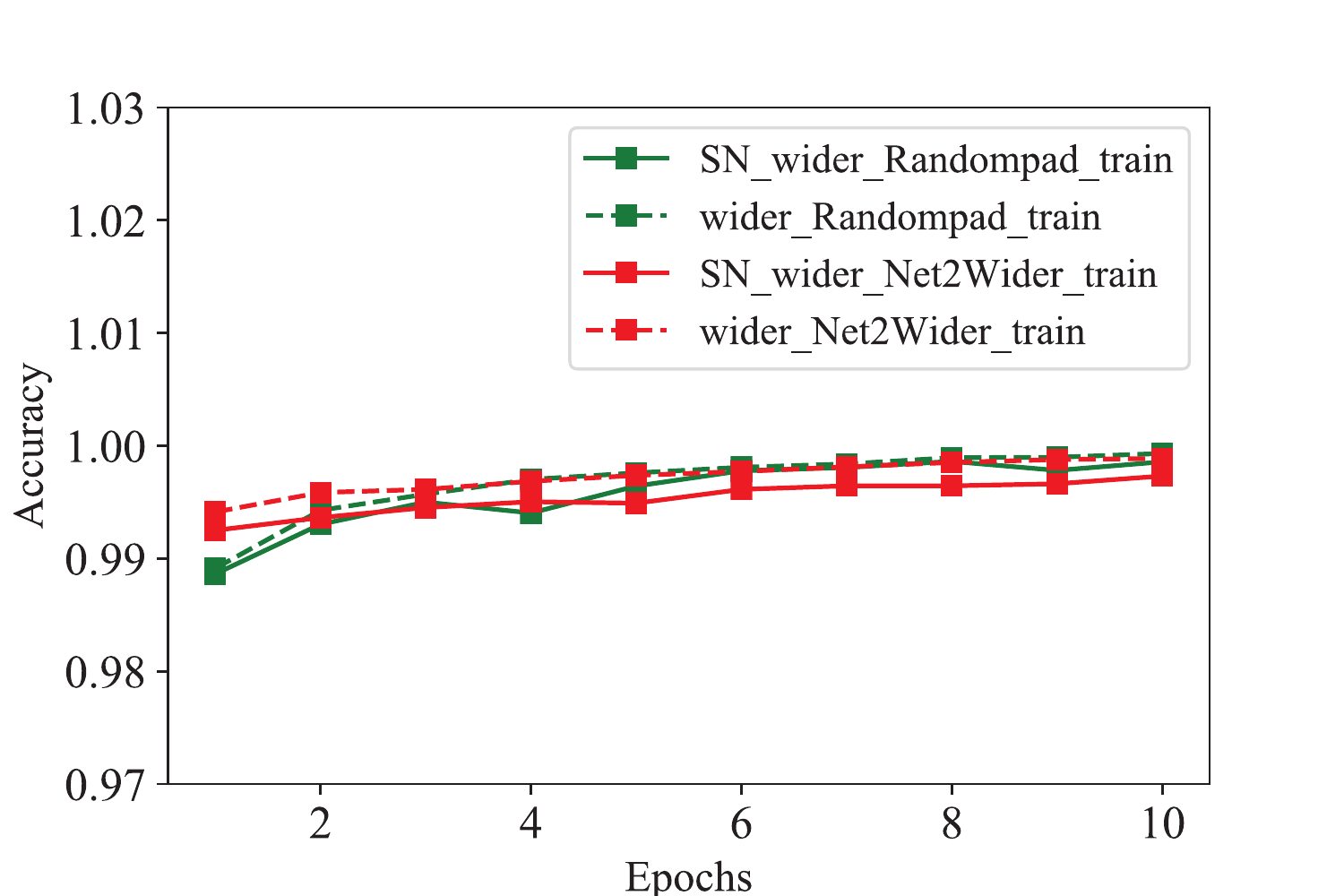}
\end{minipage}%
}%
\subfigure[deeper student\,model]{
\begin{minipage}[t]{0.3\linewidth}
\centering
\includegraphics[width=2.3in]{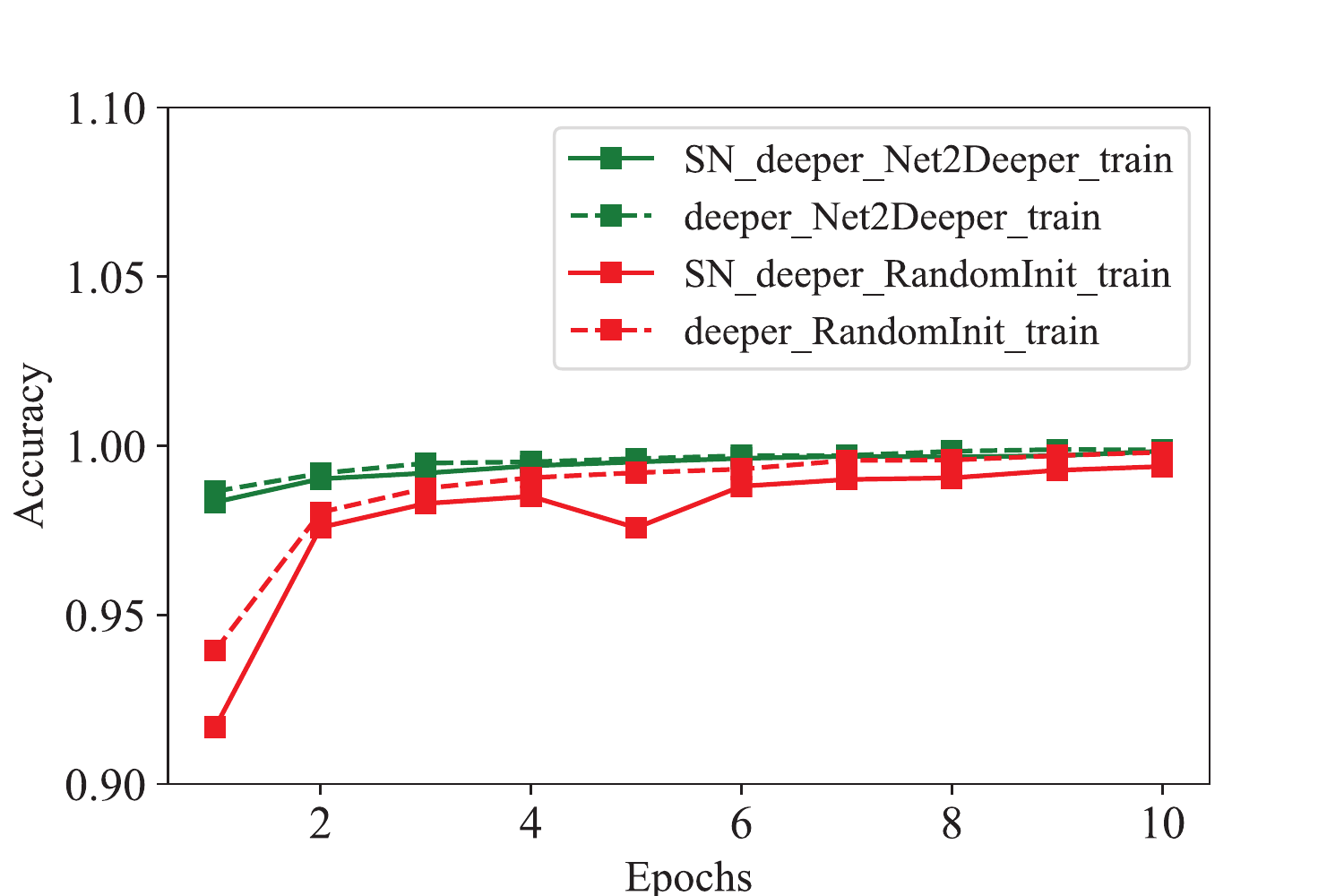}
\end{minipage}
}%
\centering
\caption{$original\,task$ accuracy over training procedure }
\label{fig4}
\end{figure*}

\begin{figure*}[htbp]
\centering
\subfigure[teacher\,model]{
\begin{minipage}[t]{0.3\linewidth}
\centering
\includegraphics[width=2.3in]{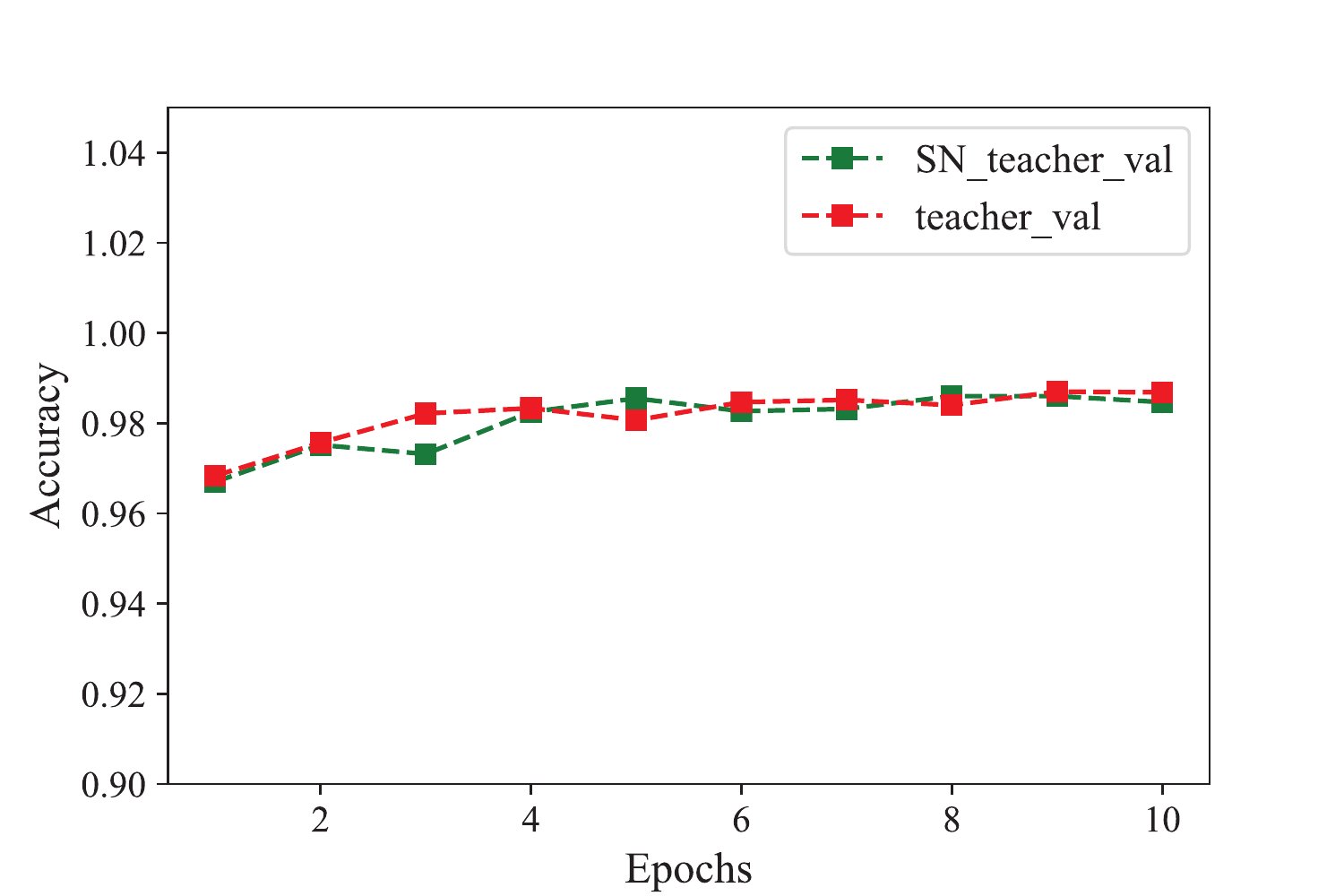}
\end{minipage}%
}%
\subfigure[wider student\,model]{
\begin{minipage}[t]{0.3\linewidth}
\centering
\includegraphics[width=2.3in]{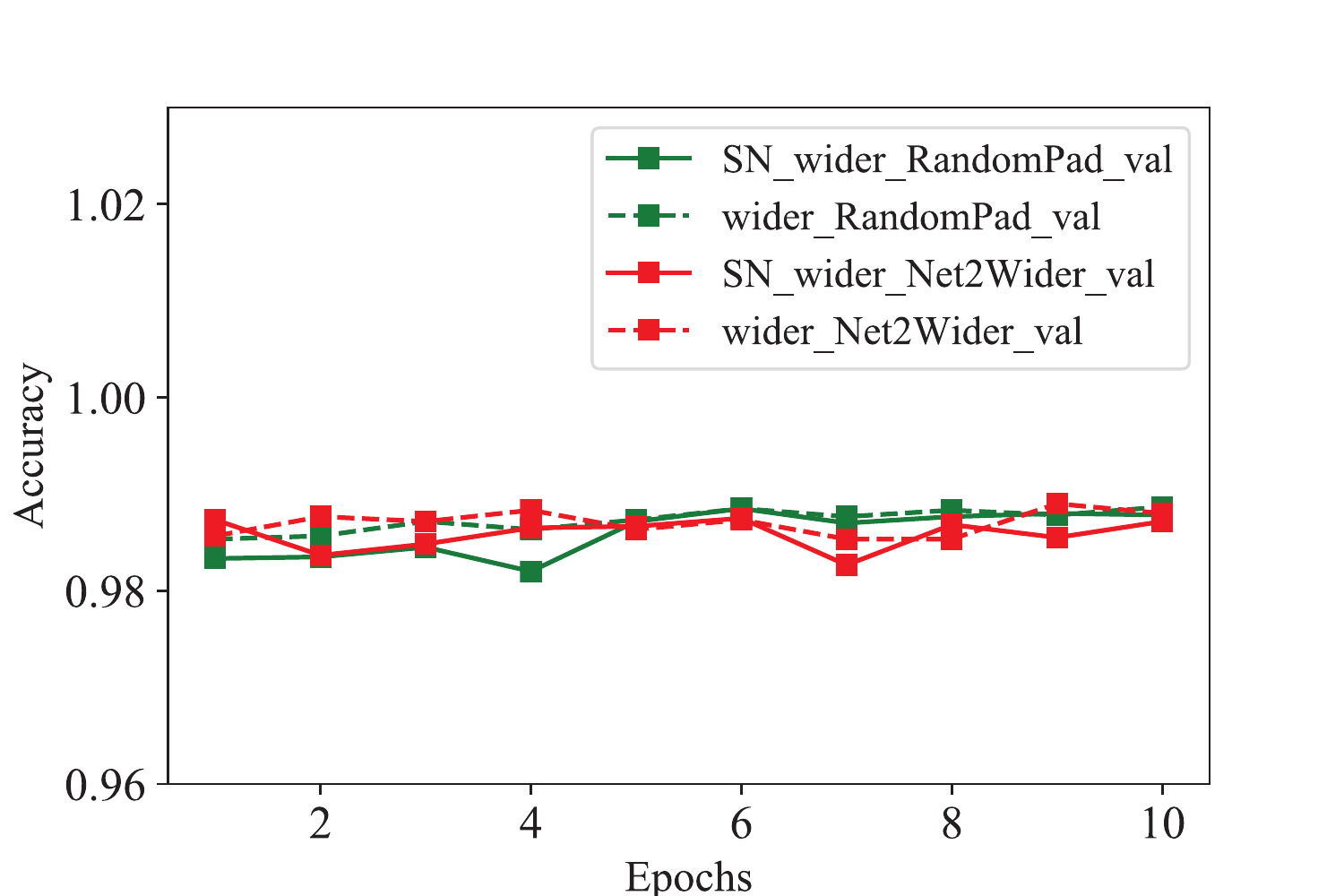}
\end{minipage}%
}%
\subfigure[deeper student\,model]{
\begin{minipage}[t]{0.3\linewidth}
\centering
\includegraphics[width=2.3in]{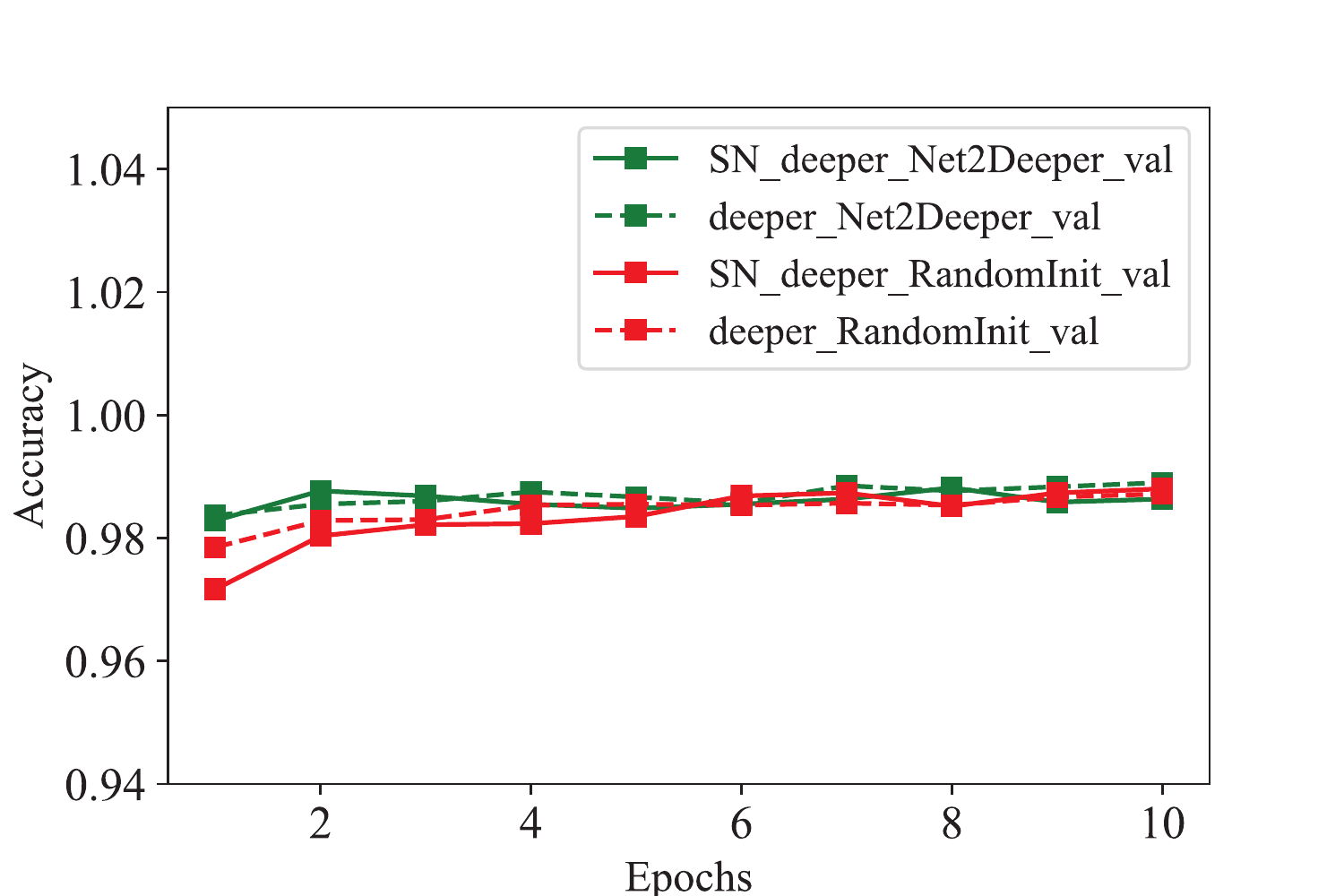}
\end{minipage}
}%
\centering
\caption{$original\,task$ accuracy in validation }
\label{fig5}
\end{figure*}

\begin{table*}
\centering
\caption{Test accuracy of different models }
\renewcommand\arraystretch{1.3}
\setlength{\tabcolsep}{7mm}{
\begin{tabular}{|c|c|c|}

\hline
  &
original\,networks  &
SN\,networks \\

\hline
Teacher &
98.95\% &
98.90\% \\

wider\_Random Pad &
99.00\% &
98.60\% \\

wider\_Net2Wider &
99.01\% &
99.00\% \\

deeper\_Random Init &
98.61\% &
99.80\% \\

deeper\_Net2Deeper &
99.15\% &
99.10\% \\
\hline
\end{tabular}}
\label{table2}
\end{table*}

\begin{table*}
\centering
\caption{SN test accuracy in different model }
\setlength{\tabcolsep}{5mm}{
\begin{tabular}{|p{1cm}<{\centering}|p{1.3cm}<{\centering}|p{1.3cm}<{\centering}|p{1.3cm}<{\centering}|p{1.3cm}<{\centering}|p{1.3cm}<{\centering}|p{1.3cm}<{\centering}|}

\hline
\multirow{2}{*}{Sparity}&
\multicolumn{2}{c|}{Teacher\_model}&
\multicolumn{2}{c|}{Net2WiderNet}&
\multicolumn{2}{c|}{Net2DeeperNet}\\
\cline{2-7}
 &orignal\,task&SN\,task&orignal\,task&SN\,task&orignal\,task&SN\,task\\
\hline
10\%&98.70\%& 100\%&98.88\%&100\%&98.62\%&100\%\\
20\%&98.62\%& 100\%&98.77\%&100\%&98.56\%&100\%\\
30\%&98.55\%& 100\%&98.73\%&100\%&98.68\%&100\%\\
40\%&98.96\%& 100\%&98.92\%&100\%&98.47\%&100\%\\
50\%&98.88\%& 100\%&98.06\%&100\%&98.84\%&100\%\\
60\%&98.95\%& 100\%&99.07\%&100\%&98.98\%&100\%\\
70\%&98.99\%& 100\%&99.11\%&100\%&99.03\%&100\%\\
80\%&99.02\%& 100\%&99.03\%&100\%&99.15\%&100\%\\
90\%&98.75\%& 100\%&98.84\%&100\%&98.91\%&100\%\\
\hline
\end{tabular}}
\label{table3}
\end{table*}

\subsection{Fidelity}
The fidelity of the embedded SN is to test the extra training cost from the embedding process, and the side effects on the original functionality. Ideally, the SN embedding algorithm should not incur too much extra training costs and not disturb the original performance of the DNN model.
We verify the fidelity of the SN framework in terms of the training process and the testing result.
\begin{itemize}
\item \emph{Fidelity on training.}\\
We choose the model convergence speed as the evaluation standard to determine the impact of Serial Number on the model training process.
The training accuracy and validation accuracy of the classification task are recorded at each training epoch for every  $original\,networks$ and $SN\,networks$ (see Fig. 4 and Fig. 5). From this, we can observe that the accuracy curve with SN is very close to the curve of original model. All the networks show a similar convergence accuracy at the almost same epoch. Thus, we can confirm that our embedded SN does not incur additional training cost and fidelity to the original model.
\item \emph{Fidelity on functionality.}\\
The fidelity on functionality express in the test accuracy of the original classification task.
Specially, we use the testing dataset to compare the original classification accuracy of  $original\,networks$ and $SN\,networks$ .
TABLE\ref{table2} shows the comparison of testing accuracy between $original\,networks$ and $SN\,networks$. Experiments show that $SN\,networks$ have the same level of accuracy as the $original\,networks$. For example, testing accuracy for the deeper\_RandomInit network is 98.61\% while the corresponding accuracy of SN\_deeper\_RandomInit network is 98.80\%, a little higher than the $original\,networks$. Other $SN\,networks$ are slightly lower than their $original\,networks$, but the worst-case only drops by 0.41\%. It is worth noting that, the student models in Net2Net models still keep their inheritance from the teacher model, so they can be trained faster. Therefore, we can confirm that our embedded SN does not disturb the original classification performance too much.
\end{itemize}

\subsection{Robustness}
As we mentioned, the state-of-art performance of deep neural networks always come at high storage space and computing resource consumption, which limit its practical application in various hardware platforms. In order to solve these problems, some methods related to model compression and acceleration have been proposed and utilized \cite{zhu2017prune},\cite{albert2004using}.Model pruning is a way to minimize the complexity of the model and speed up the training and speculation of the model with maintaining the model original performance \cite{molchanov2016pruning}.

Weight pruning is adopted in this experiment, which eliminates unnecessary values in weight tensor, reduces the number of connections between neural network layers, and reduces the parameters involved in calculation, thus reducing the operation times. For all $SN\,networks$, We specify the ultimate target sparsity (from10\% to 90\%) and then perform the pruning plan with the configuration of the pruning structure. Then we compare the accuracy of the classification task and SN task to evaluate the robust of DNN SN. Ideally, after the model pruning, the SN task still keeps high accuracy.
TABLE\ref{table3} shows the classification task accuracy and SN task accuracy of testing data for different models. For student models, Net2WiderNet and Net2DeeperNet are used.
For the SN task, even sparsity down to 90\%, our SN networks are still with 100\% accuracy, while the accuracy of the classification task drops around 2.15\% in the worst case. Therefore, our DNN SN is robust to such pruning modifications.

\section{related work}
Watermarking algorithms have been proposed to identify the deep learning models so that protecting their intellectual property\cite{nagai2018digital}. These methods are usually divided into two steps: watermark embedding and watermark verification. In the process of embedding, the model owner hides the watermarks into the training model. During the watermark verification process, the model owner can extract their hidden watermarks to verify the model belongs.

Uchida et al. probably published the first watermarking method by embedding crafted information into the weights of convolutional neural networks. However, extracting those parameters at this technique, one needs to access the model locally and entirely. Nevertheless, it is difficult to approach the parameters of the model directly, since the leaked models always not released publicly, which makes the watermarking scheme useless. Subsequently, some remote authentication watermarking methods are proposed, such as the zero-bit watermarking algorithm and digital watermarking algorithm. They rely on adversarial training samples and tweak their decision frontiers. When model verification is needed, the protected DNN model can output the pre-defined incorrect labels of these adversarial samples to validate the model ownership. However, there are virtually infinite adversarial samples, so anyone can take any adversarial sample and claim to be the owner of the model.
In this paper,we put forth a new DNN Serial Number watermarking, which is independent of the original predictions and impossible cracked.

\section{Conclusion}
In this paper, we propose a novel method to identify the deep neural network models for intellectual property protection. The key innovation of this method is that it gets rid of the traditional trigger set watermarking that is limited to the original model outputs and reuses the intermediate information of the model output layer to get the DNN Serial Number. In this way, it is hard for attackers to crack and fake the DNN SN. Also noteworthy is that we provide a new training procession, alternate training, to realize multi-tasks with no interference. Experiment results demonstrate that the DNN SN applies to many embedding situations, like training from scratch and learning from the teacher net with different learning settings. Moreover,The DNN SN only incurs slight accuracy degradation and is robust to model pruning and tampering attacks.

\clearpage

\begin{appendices}
\section{The architecture of Net2Net}

\begin{minipage}{\textwidth}
 \begin{minipage}[t]{0.45\textwidth}
  \centering
     \makeatletter\def\@captype{table}\makeatother\caption{The architecture of Teacher model}
       \begin{tabular}{cccc}
\hline
Layer Type&	Teacher\\
\hline
Conv2D	&64filter(3*3)\\
\hline
Max Pooling2D&	2*2\\
\hline
Conv2D&	64 filter(3*3)\\
\hline
Max Pooling2D&	2*2\\
\hline
Flatten&	3136\\
\hline
Dense.ReLU&	64\\
\hline
Softmax&	10\\
\hline
	\end{tabular}
  \end{minipage}
  \begin{minipage}[t]{0.45\textwidth}
   \centering
        \makeatletter\def\@captype{table}\makeatother\caption{The architecture of wider student model}
         \begin{tabular}{cccc}
\hline
Layer Type&	wider student\\
\hline
Conv2D	&128filter(3*3)\\
\hline
Max Pooling2D&	2*2\\
\hline
Conv2D&	64 filter(3*3)\\
\hline
Max Pooling2D&	2*2\\
\hline
Flatten&	3136\\
\hline
Dense.ReLU&	128\\
\hline
Softmax&	10\\
\hline
	  \end{tabular}
   \end{minipage}
   \begin{minipage}[t]{0.45\textwidth}
   \centering
        \makeatletter\def\@captype{table}\makeatother\caption{The architecture of deeper student model}
         \begin{tabular}{cccc}
\hline
Layer Type&	deeper student\\
\hline
Conv2D	&64filter(3*3)\\
\hline
Max Pooling2D&	2*2\\
\hline
Conv2D&	64 filter(3*3)\\
\hline
Conv2D&	64 filter(3*3)\\
\hline
Max Pooling2D&	2*2\\
\hline
Flatten&	3136\\
\hline
Dense.ReLU&	64\\
\hline
Dense.ReLU&	64\\
\hline
Softmax&	10\\
\hline
	  \end{tabular}
   \end{minipage}
\end{minipage}

\end{appendices}

\EOD


\begin{thebibliography}{10}

\bibitem{he2016deep}
K.~He, X.~Zhang, S.~Ren, and J.~Sun, ``Deep residual learning for image
  recognition,'' in {\em Proceedings of the IEEE conference on computer vision
  and pattern recognition}, pp.~770--778, 2016.

\bibitem{krizhevsky2012imagenet}
A.~Krizhevsky, I.~Sutskever, and G.~E. Hinton, ``Imagenet classification with
  deep convolutional neural networks,'' in {\em Advances in neural information
  processing systems}, pp.~1097--1105, 2012.

\bibitem{graves2013speech}
A.~Graves, A.-r. Mohamed, and G.~Hinton, ``Speech recognition with deep
  recurrent neural networks,'' in {\em 2013 IEEE international conference on
  acoustics, speech and signal processing}, pp.~6645--6649, IEEE, 2013.

\bibitem{hannun2014deep}
A.~Hannun, C.~Case, J.~Casper, B.~Catanzaro, G.~Diamos, E.~Elsen, R.~Prenger,
  S.~Satheesh, S.~Sengupta, A.~Coates, {\em et~al.}, ``Deep speech: Scaling up
  end-to-end speech recognition,'' {\em arXiv preprint arXiv:1412.5567}, 2014.

\bibitem{jaderberg2019human}
M.~Jaderberg, W.~M. Czarnecki, I.~Dunning, L.~Marris, G.~Lever, A.~G.
  Castaneda, C.~Beattie, N.~C. Rabinowitz, A.~S. Morcos, A.~Ruderman, {\em
  et~al.}, ``Human-level performance in 3d multiplayer games with
  population-based reinforcement learning,'' {\em Science}, vol.~364, no.~6443,
  pp.~859--865, 2019.

\bibitem{simonyan2014very}
K.~Simonyan and A.~Zisserman, ``Very deep convolutional networks for
  large-scale image recognition,'' {\em arXiv preprint arXiv:1409.1556}, 2014.

\bibitem{uchida2017embedding}
Y.~Uchida, Y.~Nagai, S.~Sakazawa, and S.~Satoh, ``Embedding watermarks into
  deep neural networks,'' in {\em Proceedings of the 2017 ACM on International
  Conference on Multimedia Retrieval}, pp.~269--277, ACM, 2017.

\bibitem{adi2018turning}
Y.~Adi, C.~Baum, M.~Cisse, B.~Pinkas, and J.~Keshet, ``Turning your weakness
  into a strength: Watermarking deep neural networks by backdooring,'' in {\em
  27th $\{$USENIX$\}$ Security Symposium ($\{$USENIX$\}$ Security 18)},
  pp.~1615--1631, 2018.

\bibitem{zhang2018protecting}
J.~Zhang, Z.~Gu, J.~Jang, H.~Wu, M.~P. Stoecklin, H.~Huang, and I.~Molloy,
  ``Protecting intellectual property of deep neural networks with
  watermarking,'' in {\em Proceedings of the 2018 on Asia Conference on
  Computer and Communications Security}, pp.~159--172, ACM, 2018.

\bibitem{kaur2012digital}
R.~Kaur and A.~Kaur, ``Digital signature,'' in {\em 2012 International
  Conference on Computing Sciences}, pp.~295--301, IEEE, 2012.

\bibitem{fan2019rethinking}
L.~Fan, K.~W. Ng, and C.~S. Chan, ``Rethinking deep neural network ownership
  verification: Embedding passports to defeat ambiguity attacks,'' in {\em
  Advances in Neural Information Processing Systems}, pp.~4716--4725, 2019.

\bibitem{hinton2015distilling}
G.~Hinton, O.~Vinyals, and J.~Dean, ``Distilling the knowledge in a neural
  network,'' {\em arXiv preprint arXiv:1503.02531}, 2015.

\bibitem{lecun1998gradient}
Y.~LeCun, L.~Bottou, Y.~Bengio, P.~Haffner, {\em et~al.}, ``Gradient-based
  learning applied to document recognition,'' {\em Proceedings of the IEEE},
  vol.~86, no.~11, pp.~2278--2324, 1998.

\bibitem{chen2015net2net}
T.~Chen, I.~Goodfellow, and J.~Shlens, ``Net2net: Accelerating learning via
  knowledge transfer,'' {\em arXiv preprint arXiv:1511.05641}, 2015.

\bibitem{zhu2017prune}
M.~Zhu and S.~Gupta, ``To prune, or not to prune: exploring the efficacy of
  pruning for model compression,'' {\em arXiv preprint arXiv:1710.01878}, 2017.

\bibitem{albert2004using}
J.~Albert, ``Using quasi-linear diffusion to model acceleration and loss from
  wave-particle interactions,'' {\em Space Weather}, vol.~2, no.~9, pp.~1--6,
  2004.

\bibitem{molchanov2016pruning}
P.~Molchanov, S.~Tyree, T.~Karras, T.~Aila, and J.~Kautz, ``Pruning
  convolutional neural networks for resource efficient inference,'' {\em arXiv
  preprint arXiv:1611.06440}, 2016.

\bibitem{nagai2018digital}
Y.~Nagai, Y.~Uchida, S.~Sakazawa, and S.~Satoh, ``Digital watermarking for deep
  neural networks,'' {\em International Journal of Multimedia Information
  Retrieval}, vol.~7, no.~1, pp.~3--16, 2018.

\bibitem{tang2019bringing}
Y.~Tang, S.~You, C.~Xu, B.~Shi, and C.~Xu, ``Bringing giant neural networks
  down to earth with unlabeled data,'' {\em arXiv preprint arXiv:1907.06065},
  2019.

\end{thebibliography}
\end{document}